\documentclass[runningheads, orivec]{llncs}
\usepackage{graphicx}
\usepackage{amsmath,amssymb} 
\usepackage{color}

\usepackage{multirow}
\usepackage[table]{xcolor}
\usepackage{overpic}
\usepackage{wrapfig,lipsum,booktabs}
\usepackage{url}
\usepackage{algorithm}
\usepackage{algorithmic}
\usepackage{subfigure}
\usepackage{rotating}

\usepackage[breaklinks=true,colorlinks,citecolor=blue,urlcolor=blue,linkcolor=red,bookmarks=false]{hyperref}

\ifdefined \GramaCheck
  \newcommand{\CheckRmv}[1]{}
  \newcommand{\figref}[1]{Figure}%
  \newcommand{\tabref}[1]{Table}%
  \newcommand{\secref}[1]{Section}
  \newcommand{\equref}[1]{Equation}
\else
  \newcommand{\CheckRmv}[1]{#1}
  \newcommand{\figref}[1]{Fig.~\ref{#1}}%
  \newcommand{\tabref}[1]{Tab.~\ref{#1}}%
  \newcommand{\secref}[1]{Sec.~\ref{#1}}
  \newcommand{\equref}[1]{Eqn.~(\ref{#1})}
\fi
\renewcommand{\eqref}[1]{Eqn.~(\ref{#1})}

\newcommand{\myPara}[1]{\paragraph{#1.}}

\newcommand{\tabSpace}{}
\newcommand{\tabFormat}{\centering \renewcommand{\arraystretch}{0.85}}
\newcommand{\printfnsymbol}[1]{%
  \textsuperscript{\@fnsymbol{#1}}%
}
\graphicspath{{./figures/}}
\usepackage[normalem]{ulem}
\definecolor{awesome}{rgb}{1.0, 0.13, 0.32}
\definecolor{hyperref-green}{RGB}{0,150,0}
\definecolor{hyperref-blue}{RGB}{0,0,200}
\definecolor{hyperref-red}{RGB}{200,0,0}

\def\ie{\emph{i.e.,~}}

\def\etc{\emph{etc}}
\def\etal{{\em et al.}}

\usepackage[width=122mm,left=12mm,paperwidth=146mm,height=193mm,top=12mm,paperheight=217mm]{geometry}

\renewcommand{\orcidID}[1]{\href{https://orcid.org/#1}{\includegraphics[scale=0.07]{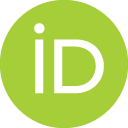}}}

\begin{document}
\pagestyle{headings}
\mainmatter
\def\ECCVSubNumber{50}  

\title{Highly Efficient Salient Object Detection with 100K Parameters} 

\titlerunning{ECCV-20 submission ID \ECCVSubNumber} 
\authorrunning{ECCV-20 submission ID \ECCVSubNumber} 
\author{Anonymous ECCV submission}
\institute{Paper ID \ECCVSubNumber}


\titlerunning{Highly Efficient Salient Object Detection with 100K Parameters}
\author{Shang-Hua Gao\inst{1}  \orcidID{0000-0002-7055-2703}\and
Yong-Qiang Tan\inst{1} \orcidID{0000-0002-0481-6715} \and 
Ming-Ming Cheng\inst{1} \orcidID{0000-0001-5550-8758} \and \\
Chengze Lu\inst{1} \orcidID{0000-0002-8225-6311} \and  
Yunpeng Chen\inst{2} \and
Shuicheng Yan\inst{2}}
\authorrunning{S. Gao et al.}
\institute{Colledge of Computer Science, Nankai University, Tianjin, China \and
  Yitu Technology \\
   \url{https://mmcheng.net/sod100k/}
}

\maketitle

\begin{abstract}
  Salient object detection models often demand a considerable 
  amount of computation cost to make precise prediction for each pixel,
  making them hardly applicable on low-power devices.
  In this paper, we aim to relieve the contradiction between computation cost and model performance by improving the network efficiency to a higher degree.
  We  propose  a  flexible  convolutional  module,  namely generalized OctConv (gOctConv),  
  to  efficiently utilize both in-stage and cross-stages multi-scale features, 
  while reducing the representation redundancy by a novel dynamic weight decay scheme.
  The effective dynamic weight decay scheme
  stably boosts the sparsity of parameters during training,
  supports learnable number of channels for each scale in gOctConv, 
  allowing $80\%$ of parameters reduce with negligible performance drop.
  Utilizing gOctConv, we build an extremely light-weighted model, namely CSNet,  which  achieves  comparable  performance  with $\sim0.2\%$  parameters (100k) of large models on popular salient object detection benchmarks.
  The source code is publicly available at \url{https://mmcheng.net/sod100k/}.
\keywords{Salient object detection, Highly efficient}
\end{abstract}

\section{Introduction}
Salient object detection (SOD) is an important computer vision task with 
various applications in image retrieval~\cite{He2012Mobile,ChengSurveyVM2017}, 
visual tracking~\cite{HongOnline},
photographic composition~\cite{han2020deep},
image quality assessment~\cite{wang2019no}, 
and weakly supervised semantic segmentation~\cite{hou2018selferasing}. 
While convolutional neural networks (CNNs) based SOD methods have made 
significant progress,
most of these methods focus on improving the state-of-the-art (SOTA) 
performance,
by utilizing both fine details and global semantics
\cite{Wang2015Deep,zhang2017learning,Zhang_2018_CVPR,Zeng_2019_ICCV,Fan2020S4Net}, 
attention~\cite{Chen2018Reverse,BorjiCVM2019},
as well as edge cues~\cite{Feng_2019_CVPR,Wang_2019_CVPR,Zhaoedge_2019_ICCV,VecRoad_20CVPR} \etc.
Despite the great performance, these models are usually resource-hungry,
which are hardly applicable on low-power devices with limited 
storage/computational capability.
How to build an extremely light-weighted SOD model with SOTA performance
is an important but less explored area.

The SOD task requires generating accurate prediction scores for every 
image pixel, 
thus requires both large scale high level feature representations for
correctly locating the salient objects,
as well as fine detailed low level representations for precise boundary
refinement~\cite{Feng_2019_CVPR,Wang_2018_CVPR,HouPami19Dss}.
There are two major challenges towards building an extremely light-weighted
SOD models.
\textbf{Firstly}, serious redundancy could appear when the low frequency nature
of high level feature meets the high output resolution of saliency maps.
\textbf{Secondly}, SOTA SOD models
\cite{Liu19PoolNet,Wu_2019_CVPR,Feng_2019_CVPR,zhang2020gicd,Liu_2018_CVPR,fan2020bbs}
usually rely on ImageNet pre-trained backbone architectures~\cite{he2016deep,gao2019res2net} to extract features,
which by itself is resource-hungry.

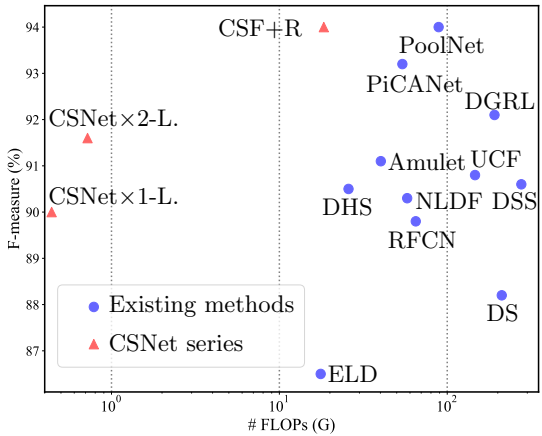
\begin{figure}[t]
  \centering
  \small
  \begin{overpic}[width=0.6\linewidth]{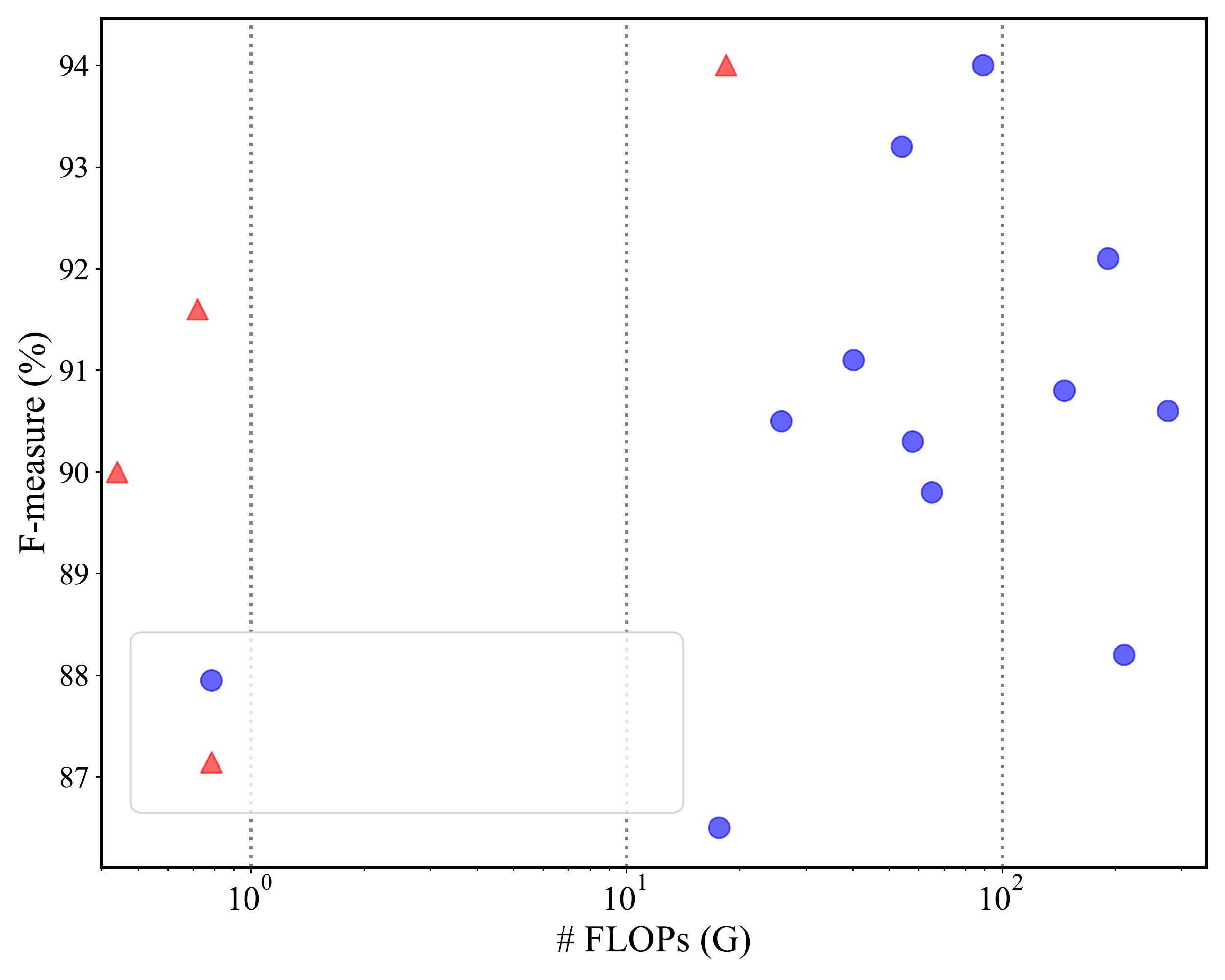}
  \put(71, 35){RFCN}
  \put(76, 42){NLDF}
  \put(90, 42){DSS}
  \put(59, 41){DHS}
  \put(71, 49){Amulet}
  \put(86, 49.5){UCF}
  \put(89, 22){DS}
  \put(60, 11){ELD}
  \put(85, 60){DGRL}
  \put(67, 63){PiCANet}
  \put(73, 70){PoolNet}
  \put(40, 73){CSF+R}
  \put(9, 43){CSNet$\times$1-L.}
  \put(9,57){CSNet$\times$2-L.}
  \put(20,23.5){Existing methods}
  \put(20,16.5){CSNet series}
  \end{overpic}
  \caption{FLOPs and performance of models on salient object detection task. 
  }\label{fig:flop_cmp_all}
\end{figure}

Very recently, the spatial redundancy issue of low frequency features 
has also been noticed by Chen \etal~\cite{chen2019drop}
in the context of image and video classification.
As a replacement of vanilla convolution, 
they design an OctConv operation to process feature maps
that vary spatially slower at a lower spatial resolution to reduce 
computational cost.
However, directly using OctConv \cite{chen2019drop} 
to reduce redundancy issue in the SOD task still faces two major challenges.
1) Only utilizing two scales, \ie low and high resolutions as in OctConv, 
is not sufficient to fully reduce redundancy issues in the SOD task, 
which needs much stronger multi-scale representation ability than 
classification tasks. 
2) The number of channels for each scale in OctConv is manually selected,   
requiring lots of efforts to re-adjust for saliency model as
SOD task requires less category information.

In this paper, we propose a generalized OctConv (gOctConv) 
for building an extremely light-weighted SOD model,
by extending the OctConv in the following aspects:
1). The flexibility to take inputs from arbitrary number of scales, 
from both in-stage features as well as cross-stages features,
allows a much larger range of multi-scale representations.
2). We propose a dynamic weight decay scheme to support 
learnable number of channels for each scale,
allowing 80$\%$ of parameters reduce with negligible performance drop.

Benefiting from the flexibility and efficiency of gOctConv,
we propose a highly light-weighted model, namely CSNet, 
that fully explores both in-stage and \textbf{C}ross-\textbf{S}tages 
multi-scale features.
As a bonus to the extremely low number of parameters,
our CSNet could be directly trained from scratch without ImageNet pre-training,
avoiding the unnecessary feature representations for distinguishing between 
various categories in the recognition task.
In summary, we make two major contributions in this paper:
\begin{itemize}
  \item We propose a flexible convolutional module, namely gOctConv,
    to efficiently utilize both in-stage and cross-stages 
    multi-scale features for SOD task, 
    while reducing the representation redundancy by a novel dynamic weight 
    decay scheme.
  \item Utilizing gOctConv, we build an extremely light-weighted SOD model,
    namely CSNet,
    which achieves comparable performance with $\sim0.2\%$ parameters (100k) 
    of SOTA large models on popular SOD benchmarks.
\end{itemize}


\section{Related Works}

\subsection{Salient Object Detection}

Early works~\cite{itti1998model,jiang2013salient,yang2013saliency,zhu2014saliency,cheng2015global} 
mainly rely on hand-craft features to detect salient objects.
\cite{Li_2016_CVPR,wang2016saliency,Liu_2016_CVPR} utilize CNNs 
to extract more informative features from image patches 
to improve the quality of saliency maps. 
Inspired by the fully convolutional networks (FCNs)~\cite{long2015fully}, 
recent works~\cite{Fan_2018_ECCV,Li_2017_CVPR,zhang2017learning,Wang_2018_CVPR,Liu_2018_CVPR,Zhang_2018_CVPR_BID}
formulate the salient object detection as a pixel-level prediction task
and predict the saliency map in an end-to-end manner using FCN based models.
\cite{HouPami19Dss,Wang_2017_ICCV,zhang2017learning,Zhang_2018_CVPR,Piao_2019_ICCV}
capture both fine details and global semantics from different stages of the backbone network.
\cite{Luo_2017_CVPR,li2018contour,Wang_2019_CVPR,Zhaoedge_2019_ICCV} introduce edge cues 
to further refine the boundary of saliency maps.
\cite{zhang2017learning,Zhao_2019_ICCV,Wu_2019_CVPR_MS} improve the saliency detection 
from the perspective of network optimization.
Despite the impressive performance, 
all these CNN based methods require ImageNet pre-trained powerful backbone networks 
as the feature extractor, which usually results in high computational cost.

\subsection{Light-weighted Models}

Currently, most light-weighted models that are initially designed for classification tasks utilize modules such as
inverted block~\cite{howard2017mobilenets,howard2019searching}, channel shuffling~\cite{zhang2018shufflenet,ma2018shufflenet}, 
and SE attention module~\cite{howard2019searching,tan2019efficientnet} to improve network efficiency.
Classification tasks~\cite{russakovsky2015imagenet} predict semantic labels for an image,
requiring more global information but fewer details.
Thus, light-weighted models
\cite{howard2017mobilenets,ma2018shufflenet,zhang2018shufflenet,howard2019searching,zhang2020split} 
designed for classification use aggressive downsampling strategies
at earlier stages to save FLOPs,
which are not applicable to be the feature extractor 
for SOD task that requires multi-scale information with
both coarse and fine features.
Also, SOD task focuses on determine the salient region
while classification tasks predicts category information.
To improve performance under limited computing budget,
the allocation of computational resources, \ie feature resolution, channels, for saliency models
should be reconsidered.

\subsection{Network Pruning}

Many network pruning methods
\cite{li2016pruning,luo2017thinet,liu2017learning,he2017channel,liu2019metapruning,he2019filter} 
have been proposed to prune unimportant filters especially on channel level.
\cite{li2016pruning,he2018soft} use the norm criterion to estimate redundant filters.
\cite{luo2017thinet} prunes filters based on statistics information of the next layer.
\cite{liu2017learning} reuses the scaling factor of BatchNorm layer
as the indicator of filter importance.
\cite{he2019filter} computes the geometric median of weights to select filters.
\cite{liu2019metapruning} utilizes generated weights to estimate the performance 
of remaining filters.
Mostly pruning approaches rely on regularization tricks such as weight decay
to introduce sparsity to filters.
Our proposed dynamic weight decay stably introduces sparsity for
assisting pruning algorithms to prune redundant filters,
resulting in learnable channels for each scale in our proposed gOctConv.


\newcommand{\addImg}[1]{\includegraphics[height=.28\linewidth]{#1}}

\begin{figure}[b]
  \centering 
  \addImg{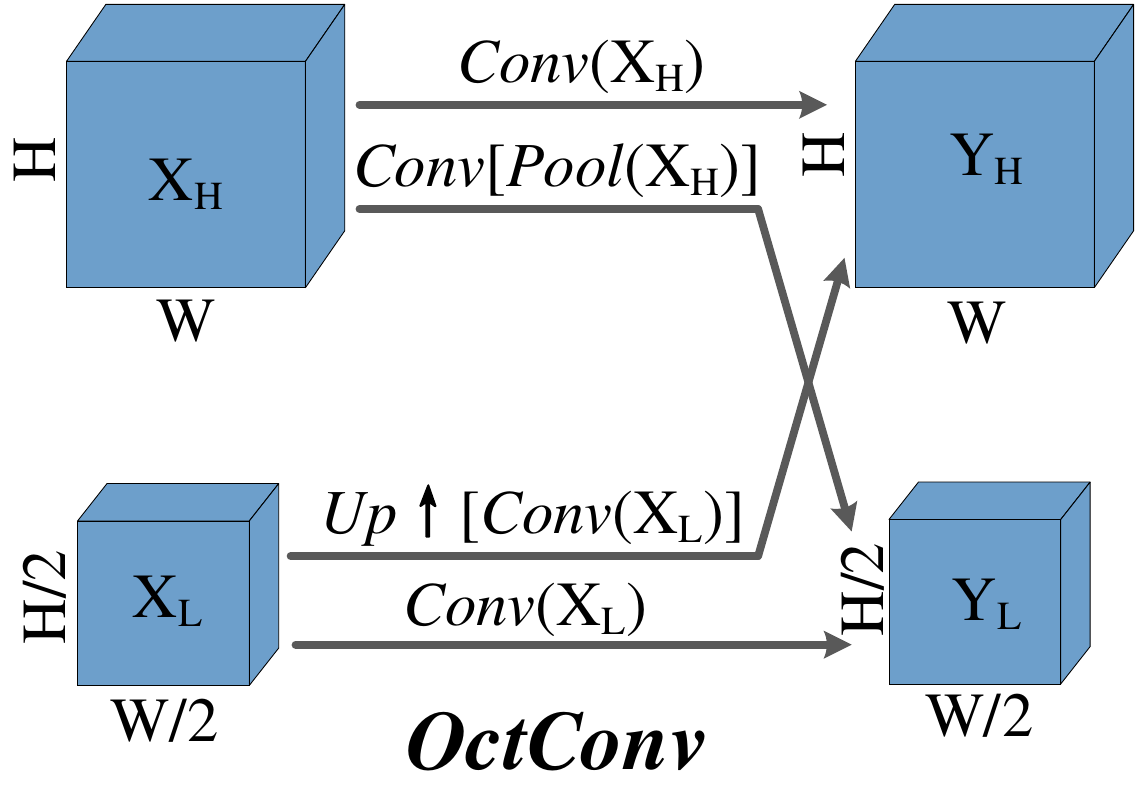} \hspace{0.4in}
  \addImg{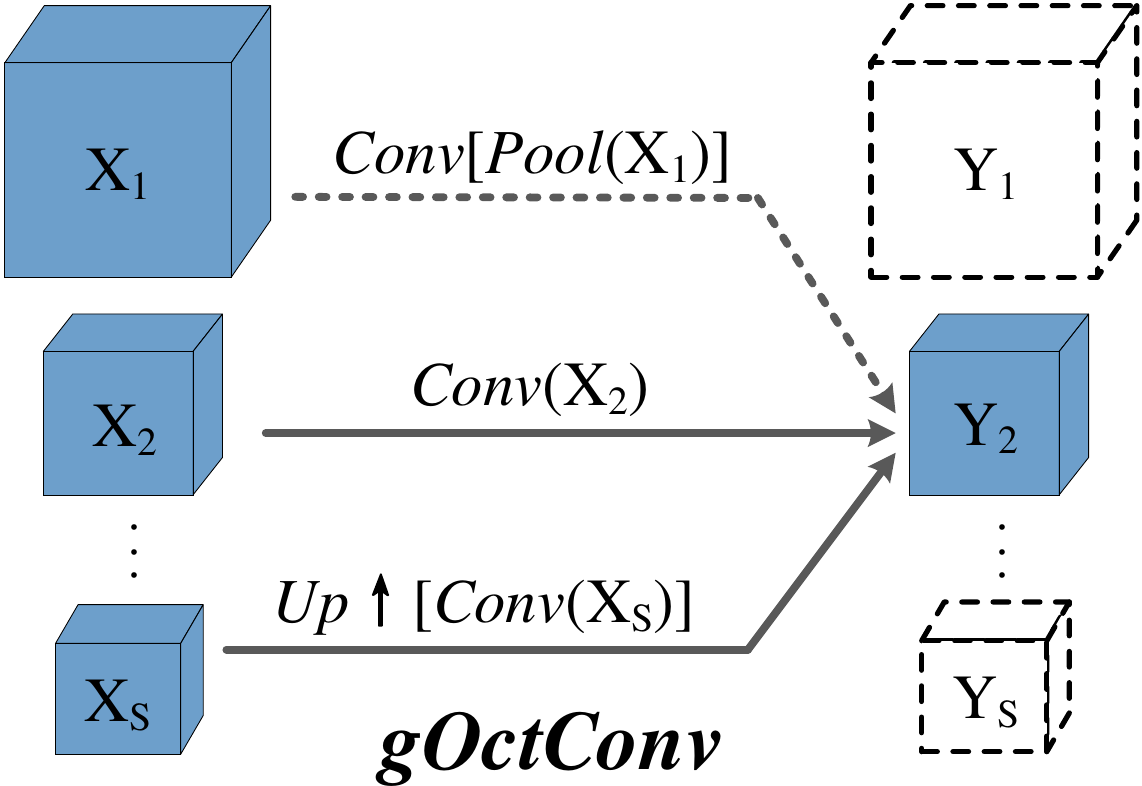} 
  \caption{While originally designed to be a replacement of 
    traditional convolution unit, 
    the OctConv~\cite{chen2019drop} takes two high/low resolution inputs 
    from the same stage with fixed number of feature channels.
    Our gOctConv allows arbitrary number of input resolutions from both
    in-stage and cross-stages conv features with learnable number of channels.
  }\label{fig:octconvs}
\end{figure}

\section{Light-weighted Network with Generalized OctConv}
\subsection{Overview of Generalized OctConv}
Originally designed to be a replacement of traditional convolution unit, the vanilla OctConv~\cite{chen2019drop} shown in ~\figref{fig:octconvs} (a) conducts the convolution operation across high/low scales within a stage.
%
However, only two-scales within a stage can not introduce enough multi-scale information required for SOD task.
The channels for each scale in vanilla OctConv is manually set, 
requires lots of efforts to re-adjust for saliency model as
SOD task requires less category information.
Therefore, we propose a generalized OctConv (gOctConv) allows arbitrary number of input resolutions from both in-stage and cross-stages conv features with learnable number of channels as shown in ~\figref{fig:octconvs} (b).
As a generalized version of vanilla OctConv, gOctConv improves the vanilla OctConv for SOD task in following aspects:
1). Arbitrary numbers of input and output scales is available to support
a larger range of multi-scale representation.
2). Except for in-stage features, the gOctConv can also process cross-stages features with arbitrary scales from the feature extractor.
3). The gOctConv supports learnable channels for each scale through our proposed dynamic weight decay assisting pruning scheme.
4). Cross-scales feature interaction can be turned off to support a large complexity flexibility.
The flexible gOctConv allows many instances under different designing requirements.  
We will give a detailed introduction of different instances of gOctConvs
in following light-weighted model designing.

\subsection{Light-weighted Model Composed of gOctConvs}
\begin{figure}[b]
  \centering
  \includegraphics[width=\linewidth]{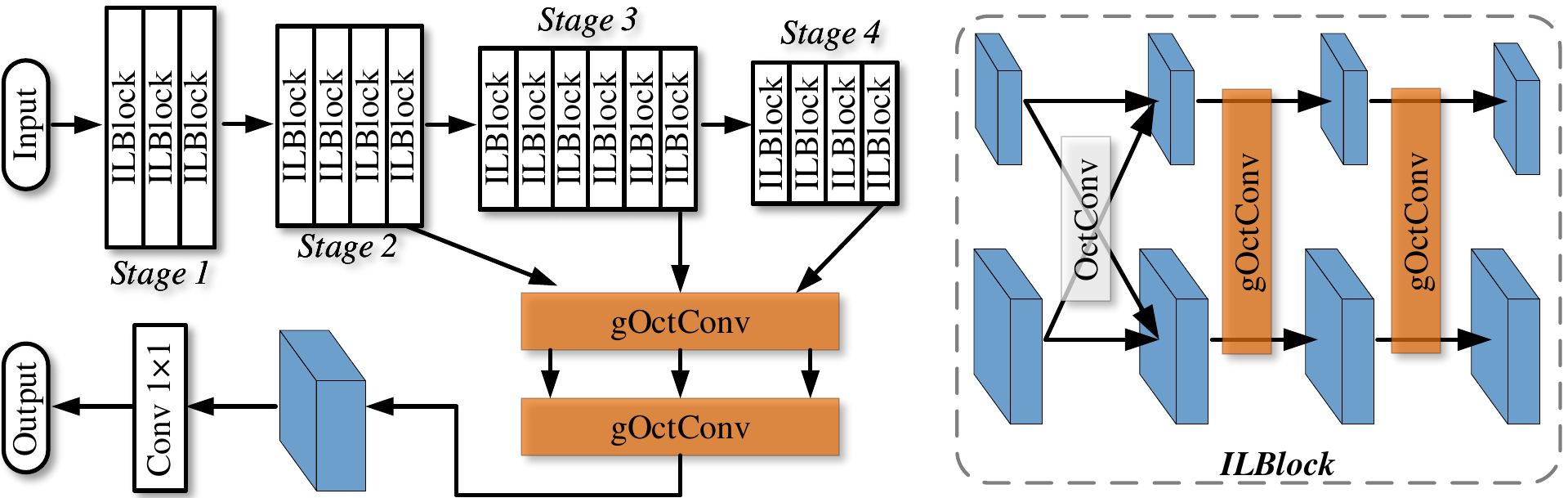}
  \caption{Illustration of our salient object detection pipeline, 
    which uses gOctConv to extract both in-stage and cross-stages
    multi-scale features in a highly efficient way.
  }\label{fig:csnet}
\end{figure}

\noindent\textbf{Overview.}As shown in \figref{fig:csnet}, 
our proposed light-weighted network, consisting of a feature extractor and a cross-stages fusion part, synchronously processes features with multiple scales.
The feature extractor is stacked with our proposed in-layer multi-scale block, namely ILBlocks,
and is split into 4 stages according to the resolution of feature maps, where each stage has 3, 4, 6, and 4 ILBlocks, respectively.
The cross-stages fusion part, composed of gOctConvs,  processes features from stages of 
the feature extractor to get a high-resolution output.

\noindent\textbf{In-layer Multi-scale Block.} 
ILBlock enhances the multi-scale representation of features within a stage.
gOctConvs are utilized to introduce multi-scale within ILBlock.
Vanilla OctConv requires about $60\%$ FLOPs~\cite{chen2019drop} to
 achieves the similar performance to standard convolution, 
which is not enough for our objective of designing a highly light-weighted model.
To save computational cost, 
interacting features with different scales in every layer is unnecessary.
Therefore, we apply an instance of gOctConv 
that each input channel corresponds to an output channel with the same resolution through eliminating the cross scale operations.
A depthwise operation within each scale in utilized to further save computational cost.
This instance of gOctConv only requires about $1/channel$ FLOPs compared
with vanilla OctConv.
ILBlock is composed of a vanilla OctConv
and two 3 $\times$ 3 gOctConvs as shown in~\figref{fig:csnet}.
Vanilla OctConv interacts features with two scales and 
gOctConvs extract features within each scale.
Multi-scale features within a block are separately processed and interacted alternately.
Each convolution is followed by the BatchNorm~\cite{ioffe2015batch} and PRelu~\cite{he2015delving}.
Initially, we roughly double the channels of ILBlocks as the resolution decreases, 
except for the last two stages that have the same channel number.
Unless otherwise stated, the channels for different scales in ILBlocks are set evenly.
Learnable channels of OctConvs then are obtained through the scheme as described in~\secref{sec:searched_octconv}.

\noindent\textbf{Cross-stages Fusion.}
To retain a high output resolution, common methods
retain high feature resolution on high-level of the feature extractor 
and construct complex multi-level aggregation module,
inevitably increase the computational redundancy.
While the value of multi-level aggregation is widely recognized~\cite{hariharan2015hypercolumns,lin2017feature}, 
how to efficiently and concisely achieve it remains challenging. 
Instead, we simply use gOctConvs
to fuse multi-scale features from stages of the feature extractor and generate the high-resolution output.
As a trade-off between efficiency and performance, features from last three stages are
used.
A gOctConv 1 $\times$ 1 takes features with different scales from the last conv of each stage as input 
and conducts a cross-stages convolution to output features with different scales. 
To extract multi-scale features at a granular level, each scale of features 
is processed by a group of parallel convolutions with different dilation rates.
Features are then sent to another gOctConv 1 $\times$ 1 to generate features with the highest resolution.
Another standard conv 1 $\times$ 1 outputs the prediction result of saliency map.
Learnable channels of gOctConvs in this part are also obtained.

\subsection{Learnable Channels for gOctConv}
We propose to get learnable channels for each scale in gOctConv
by utilizing our proposed dynamic weight decay assisted pruning during training.
Dynamic weight decay maintains a stable weights distribution among channels while introducing sparsity, 
helping pruning algorithms to eliminate redundant channels with negligible
performance drop.

\begin{figure}[t]
  \centering
  \begin{minipage}[t]{0.49\linewidth}
  \includegraphics[width=\linewidth]{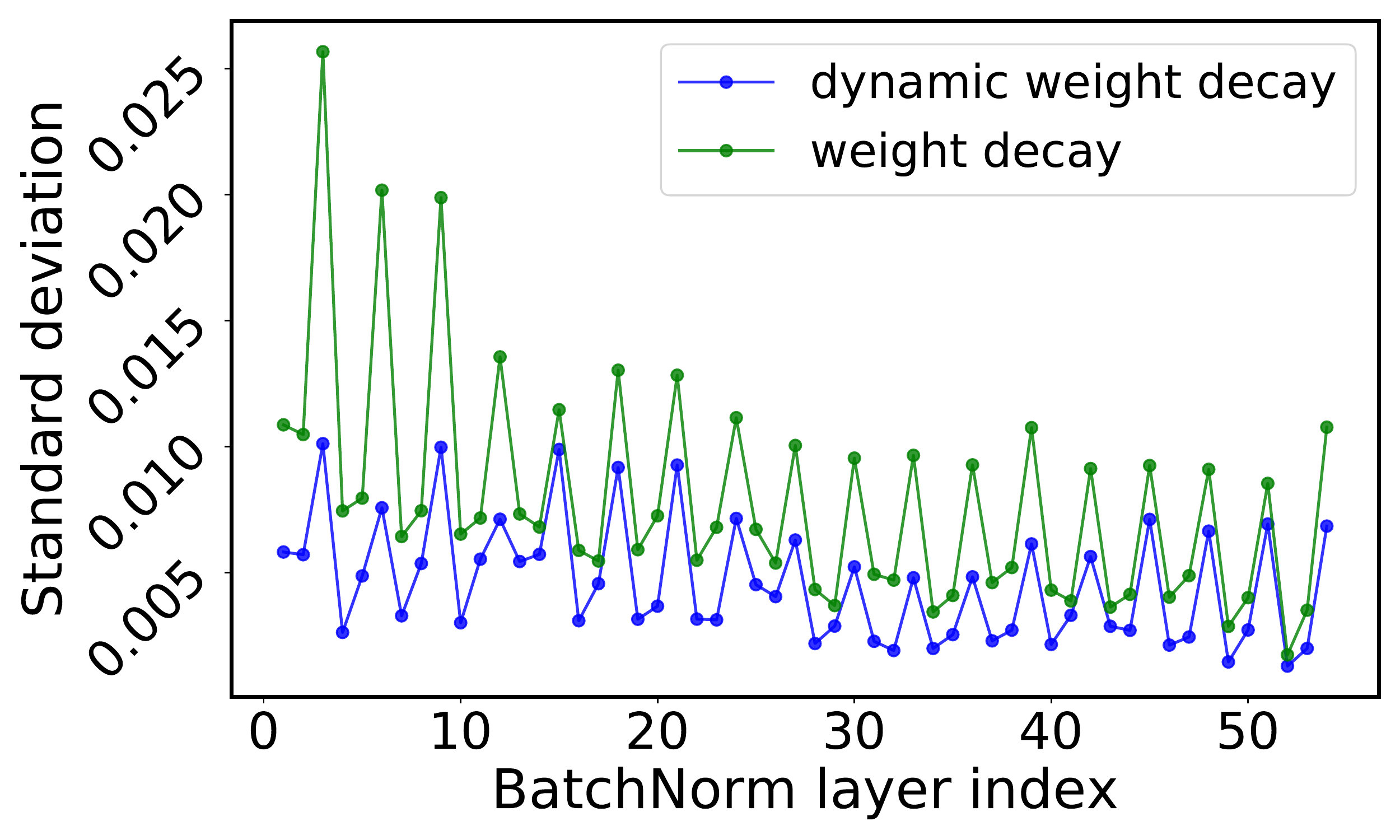}
  \centering
  \end{minipage}
  \begin{minipage}[t]{0.49\linewidth}
  \centering
  \includegraphics[width=1.\linewidth]{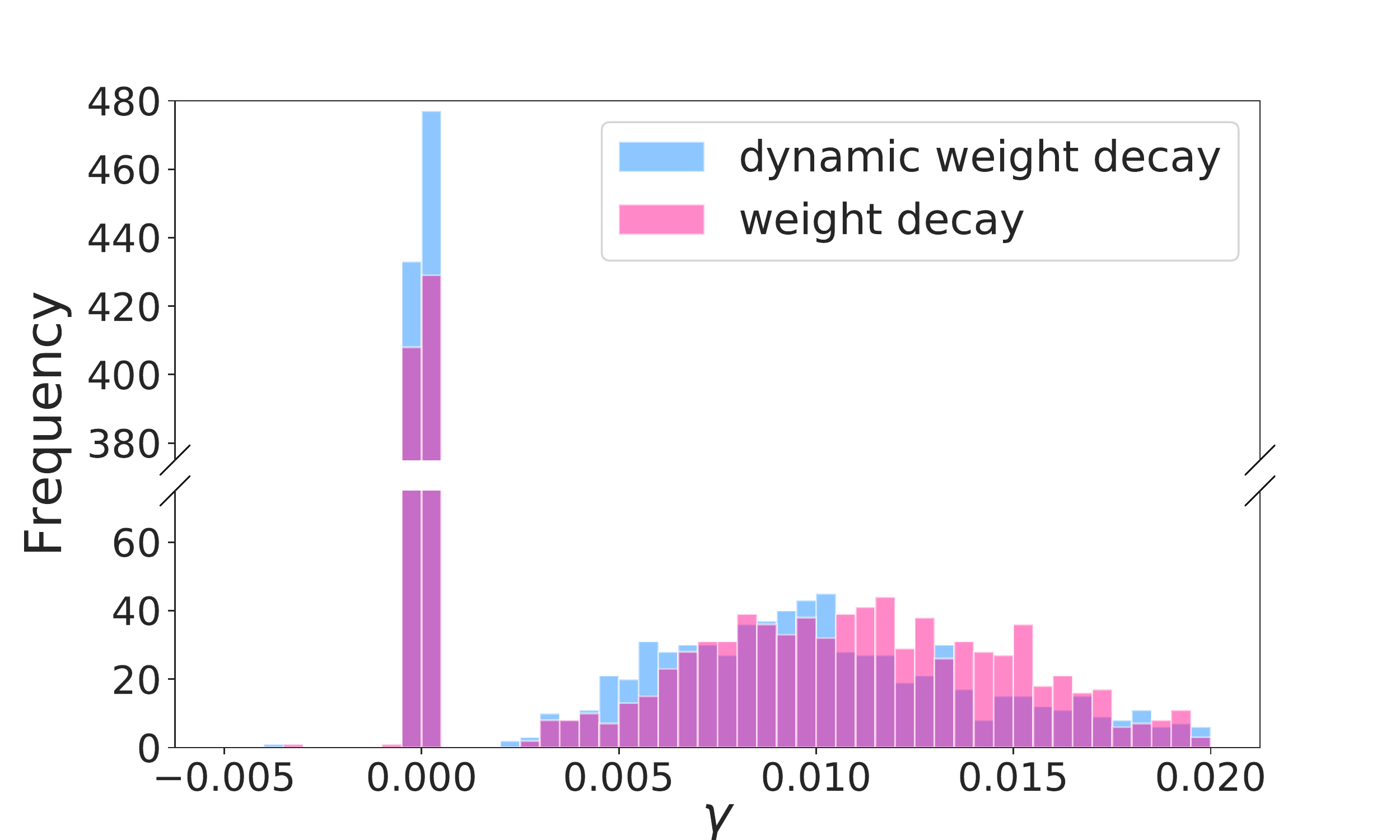}
  \end{minipage}
   \caption{a) Left: The averaged standard deviation of outputs among channels 
  from BatchNorm layer in models trained with/without dynamic weight decay. b) Right: Distribution of $\gamma$ in \eqref{eq:BatchNorm} of models 
  trained with/without dynamic weight decay.
  }\label{fig:gap_std_bn_weight_distribution}
\end{figure}

\noindent\textbf{Dynamic Weight Decay.}
The commonly used regularization trick weight decay~\cite{krogh1992simple,zhang2018three} 
endows CNNs with better generalization performance.
Mehta \etal \cite{mehta2019implicit} shows that weight decay introduces sparsity into CNNs,
which helps to prune unimportant weights.
Training with weight decay makes unimportant weights in CNN have values close to zero.
Thus, weight decay has been widely used in pruning algorithms to introduce sparsity
\cite{li2016pruning,luo2017thinet,liu2017learning,he2017channel,liu2019metapruning,he2019filter}.
The common implementation of weight decay is by adding the L2 regularization to the loss function,
which can be written as follows:
\begin{eqnarray}
  \begin{aligned}
    \boldsymbol{L} = \boldsymbol{L_0} + \lambda \sum \frac{1}{2}\boldsymbol{w_i}^2,
  \end{aligned}
  \label{eq:weight_decay}
\end{eqnarray}
where $\boldsymbol{L_0}$ is the loss for the specific task, $\boldsymbol{w_i}$ is the 
weight of $i$th layer,
and $\lambda$ is the weight for weight decay. 
During back propagation, the weight $\boldsymbol{w_i}$ is updated as
\begin{eqnarray}
  \begin{aligned}
    \boldsymbol{w_i} \leftarrow \boldsymbol{w_i}- 
    \nabla f_{i}\left(\boldsymbol{w_i}\right)  -\lambda \boldsymbol{w_i}, 
  \end{aligned}
  \label{eq:weight_decay_backprop}
\end{eqnarray}
where $\nabla f_{i}\left(\boldsymbol{w_i}\right)$ is the gradient to be updated,
and $\lambda \boldsymbol{w_i}$ is the decay term,
which is only associated with the weight itself.
Applying a large decay term enhances sparsity, 
and meanwhile inevitably enlarges the diversity of weights among channels. 
\figref{fig:gap_std_bn_weight_distribution} (a) shows that diverse weights cause unstable distribution of outputs among channels.
Ruan \etal \cite{dongsheng2019linear} reveals that channels with diverse outputs 
are more likely to contain noise, leading to biased representation for subsequent filters.  
Attention mechanisms have been widely used to re-calibrate the diverse outputs
with extra blocks and computational cost~\cite{hu2018senet,dongsheng2019linear}.
We propose to relieve diverse outputs among channels 
with no extra cost during inference.
We argue that the diverse outputs are mainly caused by the
indiscriminate suppression of decay terms to weights. 
Therefore, we propose to adjust the weight decay based on specific features of certain channels. 
Specifically, during back propagation, 
decay terms are dynamically changed according to features of certain channels.
The weight update of the proposed dynamic weight decay is written as
\begin{eqnarray}
  \begin{aligned}
    \boldsymbol{w_i} \leftarrow \boldsymbol{w_i}- \nabla f_{i}\left(\boldsymbol{w_i}\right)  
    -\lambda_{d} \operatorname{S} \left( \boldsymbol{x_i} \right) \boldsymbol{w_i}, 
  \end{aligned}
  \label{eq:dynamic_weight_decay_backprop}
\end{eqnarray}
where $\lambda_{d}$ is the weight of dynamic weight decay,
$\boldsymbol{x_i}$ denotes the features calculated by $\boldsymbol{w_i}$, 
and $\operatorname{S} \left( \boldsymbol{x_i} \right)$ is the metric of the feature,
which can have multiple definitions depending on the task.
In this paper, our goal is to stabilize the weight distribution
among channels according to features.
Thus, we simply use the global average pooling (GAP)~\cite{lin2013network} as the metric
for a certain channel:
\begin{eqnarray}
  \begin{aligned}
    \operatorname{S}\left( \boldsymbol{x_i} \right) = \frac{1}{HW} \sum_{h=0}^{H} 
    \sum_{w=0}^{W} \boldsymbol{x_i}_{h,w},
  \end{aligned}
  \label{eq:importance_metric}
\end{eqnarray}
where $H$ and $W$ are the height and width of the feature map $\boldsymbol{x_i}$.
The dynamic weight decay with the GAP metric ensures that 
the weights producing large value features are suppressed,
giving a compact and stable weights and outputs distribution as revealed 
in~\figref{fig:gap_std_bn_weight_distribution}.
Also, the metric can be defined as other forms to suit certain tasks
as we will study in our future work.
Please refer to~\secref{sec:dynamic_wd_exp} for a more detailed interpretation of dynamic weight decay.

\noindent\textbf{Learnable channels with model compression.} \label{sec:search_split_ratio_theory}
Now, we incorporate dynamic weight decay with pruning algorithms to
remove redundant weights, so as to get learnable channels of each scale in gOctConvs.
In this paper, we follow~\cite{liu2017learning} to use the weight of BatchNorm layer as the indicator
of the channel importance.
The BatchNorm operation~\cite{ioffe2015batch} is written as follows:
\label{sec:searched_octconv}
\begin{eqnarray}
  \begin{aligned}
    y=\frac{x-\mathrm{E}(x)}{\sqrt{\operatorname{Var}(x)+\epsilon}}  \gamma+\beta,
  \end{aligned}
  \label{eq:BatchNorm}
\end{eqnarray}
where $x$ and $y$ are input and output features, $\mathrm{E}(x)$ and $\operatorname{Var}(x)$ are the mean and variance, respectively, 
and $\epsilon$ is a small factor to avoid zero variance.
$\gamma$ and $\beta$ are learned factors.
 We apply the dynamic weight decay to $\gamma$ during training.
\figref{fig:gap_std_bn_weight_distribution} (b) reveals that there is a clear gap between important and redundant weights,
and unimportant weights are suppressed to nearly zero ($\boldsymbol{w_i}<1e-20$).
Thus, we can easily remove channels whose $\gamma$ is less than a small threshold.
The learnable channels of each resolution features in gOctConv are obtained.
The algorithm of getting learnbale channels of gOctConvs is illustrated in Alg.~\ref{alg:search_split_ratio}.
\begin{algorithm}
  \caption{Learnable Channels for gOctConv with Dynamic Weight Decay} 
  \label{alg:search_split_ratio}
  \begin{algorithmic}[1]
  \REQUIRE The initial CSNet in which channels for all scales in gOctConvs are  set.
    Input images $X$ and corresponding label $Y$. 
  \FOR{each iteration $i \in [1, Max Iteration]$}
  \STATE Feed input $X$ into the network to get the result $\hat{Y}$;
  \STATE Compute $Loss = criterion(\hat{Y}, Y)$;
  \STATE Compute metric for each channel using \eqref{eq:importance_metric};
  \STATE Backward with dynamic weight decay using \eqref{eq:dynamic_weight_decay_backprop}. 
  \ENDFOR
  \STATE Eliminate redundant channels to get learnable channels for each scale in gOctConv.
  \STATE Train for several iterations to fine-tune remaining weights.
  \end{algorithmic}
\end{algorithm}


\section{Experiments}

\subsection{Implementation}
\noindent\textbf{Settings.}
The implementation of the proposed method is based on the PyTorch framework.
For light-weighted models, we train models using the Adam optimizer~\cite{kingma2014adam} with a batch-size of 24 for 300 epochs from scratch.
Even with no ImageNet pre-training, the proposed CSNet still achieves comparable performance to
large models based on pre-trained backbones~\cite{simonyan2014very,he2016deep}.
The learning rate is set to 1e-4 initially, and divided by 10 at the epochs of 200, and 250.
We eliminate redundant weights and fine-tune the model for the last 20 epochs to  
compress models and get gOctConvs with the learnable channels of different resolutions.
We only use the data augmentation of random flip and crop.
The weight decay of BatchNorms following gOctConvs is replaced with our proposed dynamic weight decay with the weight of 3 by default while the weight decay
for other weights is set to 5e-3 by default.
For large models based on the pre-trained backbones, we train our models following the implementation of ~\cite{Liu19PoolNet}.
The commonly used evaluation metrics 
maximum F-measure ($F_\beta$)~\cite{achanta2009frequency}
and MAE ($M$)~\cite{Cheng_Saliency13iccv} are used for evaluation. 
FLOPs of light-weighted models are computed with an image size of 224 $\times$ 224.

\noindent\textbf{Datasets.} 
%
We following common settings of recent methods
\cite{Zhang_2018_CVPR,Liu_2018_CVPR,Liu19PoolNet,Wang_2018_CVPR,Wang_2017_ICCV,Zhaoedge_2019_ICCV} 
to train our models using the DUTS-TR~\cite{wang2017} dataset, and
evaluate the performance on several commonly used datasets, including ECSSD~\cite{Yan_2013_CVPR}, PASCAL-S~\cite{Li_2014_CVPR},
DUT-O~\cite{yang2013saliency}, HKU-IS~\cite{Li_2015_CVPR}, SOD~\cite{Movahedi2010Design}, and DUTS-TE~\cite{wang2017}.
On ablation studies, the performance on the ECSSD dataset is reported if not mentioned otherwise.


\begin{figure}[t]
  \centering
  \begin{overpic}[width=0.95\linewidth]{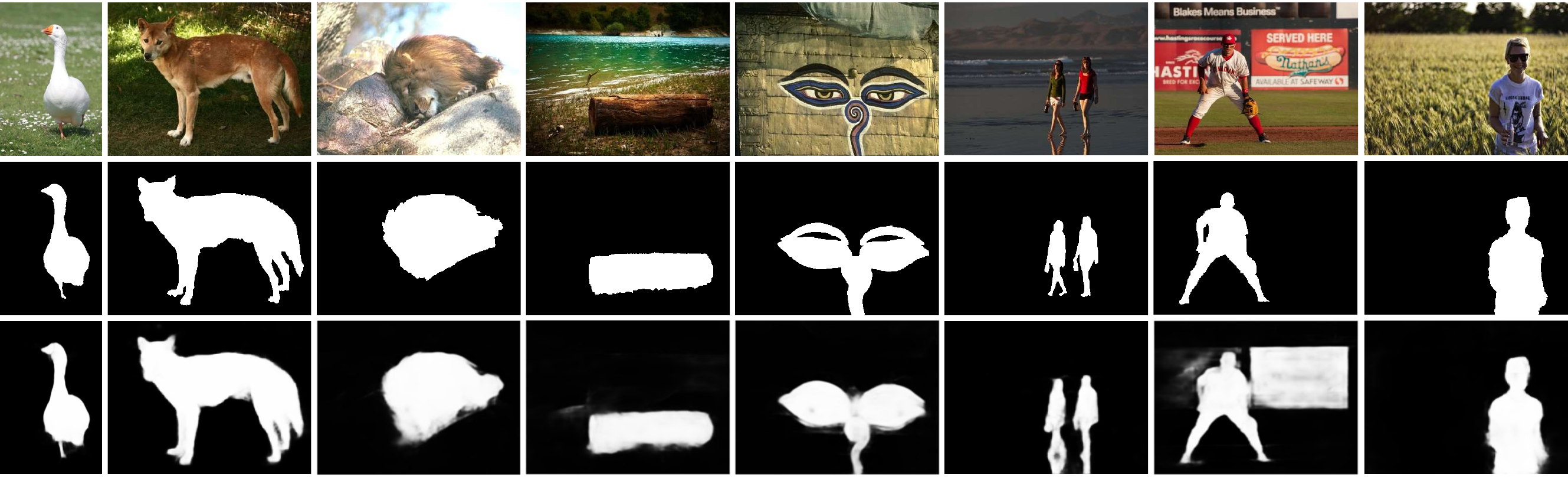}
  \end{overpic}
  \caption{Visualisation of predicted results on salient object detection. 
  Each row gives the image, GT, and predicted result, respectively.
  }\label{fig:sal_results_vis}
\end{figure}

\subsection{Performance Analysis}
\label{sec:performance}
In this section, we firstly evaluate the performance of our proposed light-weighted model CSNet with fixed channels.
Then, the performance of CSNet with learnable channels using dynamic weight decay is also evaluated.
\figref{fig:sal_results_vis} shows the visualized results of salient
object detection using our proposed light-weighted CSNet.
Also, we transfer the proposed cross-stages fusion part to commonly used large backbones~\cite{he2016deep} to verify the cross-stages feature extraction ability.

\noindent\textbf{Performance of CSNet with fixed channels in gOctConv.}
The extractor model only composed of ILBlocks.
With fixed parameters, 
we adjust the split-ratio of channels for high/low resolution features
in gOctConvs of ILBlocks
to construct models with different FLOPs, denoted by $C_H / C_L$.
\tabref{tab:performance_octhead} shows feature extraction models with different split-ratios of high/low resolution features.
Extractors achieve an low complexity thanks to
the simplified instance of gOctConvs.
Benefiting from the in-stage multi-scale representation and the low scale features in ILBlock,
 the extractor-3/1 achieves performance gain of $0.4\%$ in terms of F-measure with $80 \%$ FLOPs 
over the extractor-1/0.
The gOctConvs in cross-stages fusion part enhance the cross-stages multi-scale ability of the network while
maintaining the high output resolution by utilizing features from different stages.
As shown in~\tabref{tab:performance_octhead},
the CSNet-5/5 surpasses the extractor-3/1 by $1.4\%$ in terms of F-measure with fewer FLOPs.
Even in extreme case that the CSNet-0/1 with only low resolution features in extractor has comparable performance
 with $~44\%$ FLOPs over extractor-1/0 with all high resolution features.
However, manually tune the overall split-ratio of feature channels of different resolution
may achieves sub-optimal balance between performance and computational cost. 

To further verify the effectiveness of the cross-stage fusion (CSF) part on large models, we implement this part
into commonly used backbone network ResNet~\cite{he2016deep} and Res2Net~\cite{gao2019res2net}.
As shown in~\tabref{tab:performance_sota},
the ResNet+CSF achieves similar performance to the ResNet+PoolNet with $~53\%$ parameters and $~21\%$ FLOPs.
Unlike other models such as PoolNet that eliminates downsampling operations 
to maintain a high feature resolution on high-levels of the backbone,
the gOctConvs obtains both high and low resolution features across different stages of the backbone, 
getting a high-resolution output while saving a large amount of computational cost.
When utilizing the recently proposed Res2Net as the backbone network, the performance is further boosted.  

\newcommand{\mtd}[2]{\multirow{#1}{*}{#2}}

\begin{table}[tbp]
  \tabFormat
  \scriptsize
  \setlength{\tabcolsep}{1.4mm}
  \centering
  \begin{tabular}{lcccccc}\toprule
  Method                    &           & PARM. & FLOPs             & $F_\beta \uparrow$   & $M\downarrow$   \\ \midrule
  \mtd{5}{Extractor}& 1/0 & 180K & 0.80G & 88.2 & 0.088 \\ 
                    & 3/1 & 180K & 0.64G & \textbf{88.6} & \textbf{0.085} \\ 
                    & 5/5 & 180K & 0.45G & 88.1 & 0.086 \\ 
                    & 1/3 & 180K & 0.30G & 87.4 & 0.090 \\ 
                    & 0/1 & 180K & 0.20G & 86.4 & 0.095 \\ \midrule
  \mtd{5}{CSNet}    & 1/0 & 211K & 0.91G & 90.0 & \textbf{0.076} \\ 
                    & 3/1 & 211K & 0.78G & 89.9 & 0.077 \\ 
                    & 5/5 & 211K & 0.61G & \textbf{90.0} & 0.077 \\ 
                    & 1/3 & 211K & 0.47G & 89.2 & 0.082 \\ 
                    & 0/1 & 211K & 0.35G & 88.2 & 0.089 \\ \midrule
  \mtd{2}{CSNet-L} 
              & $\times$2 & 140K & 0.72G & \textbf{91.6} & \textbf{0.066} \\ 
              & $\times$1 & 94K  & 0.43G & 90.0 & 0.075 \\ 
\bottomrule 
\end{tabular}
\tabSpace
\caption{Performance of CSNet with the fixed split-ratio of channels in gOctConvs,
and CSNet with learnable channels.
CSNet denotes the CSNet with the fixed split-ratio in gOctConvs.
Extractor denotes the network only composed of ILBlocks. 
CSNet-L denotes the model with learnable channels using Alg.~\ref{alg:search_split_ratio}.
}
\label{tab:performance_octhead}
\end{table}

\noindent\textbf{Performance of CSNet with learnable channels in gOctConv.}
We further train the model with our proposed dynamic weight decay and 
get the learnable channels for gOctConv
as described in Alg.~\ref{alg:search_split_ratio}, named CSNet-L.
The channel for each gOctConv is expanded to enlarge the available space
for compression.
Models with channels expanded to $k$ times are denoted by CSNet-$\times k$.
\tabref{tab:compression_width} shows that our proposed dynamic weight decay
assisted pruning scheme can compress the model up to 18$\%$ of the original model size with negligible performance drop.
Compared with manually tuned split-ratio of feature resolution, 
the learnable channels of gOctConvs obtained by model compression achieves much better efficiency.
As shown in~\tabref{tab:performance_octhead}, the compressed CSNet$\times$2-L
outperforms the CSNet-5/5 by $1.6\%$ with fewer parameters and $18\%$ additional FLOPs.
The CSNet$\times$1-L achieves comparable performance compared with CSNet-5/5
with $45\%$ parameters and $70\%$ FLOPs.
\tabref{tab:performance_sota} shows that CSNet-L series achieve comparable performance 
compared with some models with extensive parameters 
such as SRM~\cite{Wang_2017_ICCV}, and Amulet~\cite{zhang2017learning} with $\sim 0.2\%$ parameters.
Note that our light-weighted models are trained from scratch while those large
models are pre-trained with ImageNet.
The performance gap between the proposed CSNet 
and the SOTA models with extensive parameters and FLOPs is only $\sim 3\%$,
considering that CSNet has the limited capacity with about $0.2\%$ parameters of large models.
We believe that more techniques such as ImageNet pre-training will bring more performance gain.

\noindent\textbf{Comparison with light-weighted models.} To the best of our knowledge,
we are the first work that aims to design an extremely light-weighted model 
for SOD task. Therefore, we port several SOTA light-weighted models
designed for other tasks such as classification and semantic segmentation for comparison.
All models share the same training configuration with our training strategy.
~\tabref{tab:performance_sota} shows that our proposed models
have considerable improvements compared with those light-weighted models.

\begin{table}[tbp]
  \tabFormat
  \scriptsize  
  \setlength{\tabcolsep}{0.55mm}
  \begin{tabular}{lcccccccccccccc}\toprule
  \multirow{2}{*}{Model} & \multicolumn{2}{c}{Complexity}  & \multicolumn{2}{c}{ECSSD} & \multicolumn{2}{c}{PASCAL-S} & \multicolumn{2}{c}{DUT-O} & \multicolumn{2}{c}{HKU-IS} & \multicolumn{2}{c}{SOD} & \multicolumn{2}{c}{DUTS-TE} \\
    & \#PARM. & FLOPs & $F_\beta$ & $M$ & $F_\beta$ & $M$ & $F_\beta$ & $M$ & $F_\beta$ & $M$ & $F_\beta$ & $M$ & $F_\beta$ & $M$ \\
  \midrule
  ELD~\cite{lee2016saliency}                     & 43.15M           & 17.63G    & .865       & .981       & .767         & .121        & .719       & .091       & .844        & .071       & .760      & .154      & -        & -        \\
  DS~\cite{DeepSaliency}                     & 134.27M           & 211.28G      & .882       & .122       & .765         & .176        & .745       & .120       & .865        & .080       & .784      & .190      & .777        & .090        \\
  DCL~\cite{Li_2016_CVPR}                     & -           & -                 & .896       & .080       & .805         & .115        & .733       & .094       & .893        & .063       & .831      & .131      & .786        & .081        \\
  RFCN~\cite{wang2016saliency}                & 19.08M      & 64.95G            & .898       & .097       & .827         & .118        & .747       & .094       & .895        & .079       & .805      & .161      & .786        & .090        \\
  DHS~\cite{Liu_2016_CVPR}                    & 93.76M      & 25.82G            & .905       & .062       & .825         & .092        & -           & -           & .892        & .052       & .823      & .128      & .815        & .065        \\
  MSR~\cite{Li_2017_CVPR}                     & -           & -                 & .903       & .059       & .839         & .083        & .790       & .073       & .907        & .043       & .841      & .111      & .824        & .062        \\
  DSS~\cite{HouPami19Dss}                     & 62.23M      & 276.37G           & .906       & .064       & .821         & .101        & .760       & .074       & .900        & .050       & .834      & .125      & .813        & .065        \\
  NLDF~\cite{Luo_2017_CVPR}                   & 35.48M      & 57.73G            & .903       & .065       & .822         & .098        & .753       & .079       & .902        & .048       & .837      & .123      & .816        & .065        \\
  UCF~\cite{zhang2017learning}                  & 29.47M      & 146.42G           & .908       & .080       & .820         & .127        & .735       & .131       & .888        & .073       & .798      & .164      & .771        & .116        \\
  Amulet~\cite{Amu_2017_ICCV}               & 33.15M      & 40.22G              & .911       & .062       & .826         & .092        & .737       & .083       & .889        & .052       & .799      & .146      & .773        & .075        \\
  GearNet~\cite{hou2018three}                 & -           & -                 & .923       & .055       & -             & -            & .790       & .068       & .934        & .034       & .853      & .117      & -            & -            \\
  PAGR~\cite{Zhang_2018_CVPR}                 & -           & -                 & .924       & .064       & .847         & .089        & .771       & .071       & .919        & .047       & -          & -          & .854        & .055        \\
  SRM~\cite{Wang_2017_ICCV}                  & 53.14M      & 36.82G             & .916       & .056       & .838         & .084        & .769       & .069       & .906        & .046       & .840      & .126      & .826        & .058        \\
  DGRL~\cite{Wang_2018_CVPR}                 & 161.74M     & 191.28G            & .921       & .043       & .844         & .072        & .774       & .062       & .910        & .036       & .843      & .103      & .828        & .049        \\
  PiCANet~\cite{Liu_2018_CVPR}               & 47.22M      & 54.05G             & .932       & .048       & .864         & .075        & .820       & .064       & .920        & .044       & .861      & .103      & .863        & .050        \\
  PoolNet~\cite{Liu19PoolNet}                & 68.26M      & 88.89G             & .940       & .042       & .863         & .075        & .830       & .055       & .934        & .032       & .867      & .100      & .886        & .040        \\ \midrule
  \multicolumn{10}{l}{Light-weighted models designed for other tasks:}\\ \midrule
  Eff.Net~\cite{tan2019efficientnet}    & 8.64M       & 2.62G                   & .828       & .129       & .739         & .158        & .696       & .129       & .807        & .116       & .712      & .199      & .687        & .135        \\     
  Sf.Netv2~\cite{ma2018shufflenet}           & 9.54M       & 4.35G              & .870       & .092       & .781         & .127        & .720       & .100       & .853        & .078       & .779      & .163      & .743        & .096        \\     
  ENet~\cite{paszke2016enet}          & 0.36M       & 0.40G                     & .857       & .107       & .770         & .138        & .730       & .109       & .839        & .094       & .741      & .183      & .730        & .111        \\     
  CGNet~\cite{wu2018cgnet}        & 0.49M       & 0.69G                         & .868       & .099       & .784         & .130        & .727       & .108       & .849        & .088       & .772      & .168      & .742        & .106        \\     
  DABNet~\cite{li2019dabnet}         & 0.75M       & 1.03G                      & .877       & .091       & .790         & .123        & .747       &  .094      & .862        & .078       & .778      & .157      & .759        & .093        \\  
  ESPNetv2~\cite{mehta2019espnetv2}          & 0.79M       & 0.31G              & .889       & .081       & .795         & .119        & .760       & .088       & .872        & .069       & .780      & .157      & .765        & .089        \\ 
   BiseNet~\cite{yu2018bisenet}  & 12.80M      & 2.50G                          & .894       & .078       & .817         & .115        & .762       &  .087      & .872        & .071       & .796      & .148      & .778        & .084        \\   \midrule
  \multicolumn{10}{l}{Ours:}\\ \midrule
   CSF+R                                  & 36.37M      & 18.40G             & .940       & .041       & .866         & .073        & .821       & .055       & .930        & .033       & .866      & .106      & .881        & .039        \\ 
  CSF+R2                              & 36.53M      & 18.96G                 & .947       & .036       &  .876 & .068 & .833 & .055 & .936 & .030 & .870 & .098 & .893 & .037              \\  \midrule
  CSNet$\times$1-L    & 94K     & 0.43G                                         & .900      & .075      & .819     & .110      & .777   & .087       & .889    & .065     & .809   & .149     & .799      & .082       \\ 
  CSNet$\times$2-L    & 140K       & 0.72G                                      & .916       & .066       & .835         & .102        & .792       & .080       & .899        & .059       & .825      & .137      & .819        & .074        \\ 
  \bottomrule 
\end{tabular}
\tabSpace
\caption{Performance and complexity comparison with state-of-the-art methods. +R, R2 denotes using the ImageNet pre-trained ResNet50~\cite{he2016deep} and Res2Net50~\cite{gao2019res2net} backbone network.
Unlike previous methods that require the ImageNet pre-trained backbone, 
our proposed light-weighted CSNet is trained from scratch without ImageNet pre-training. }
\label{tab:performance_sota}
\end{table}

\begin{figure}[t]
  \centering
  \begin{minipage}[t]{0.49\linewidth}
  \includegraphics[width=\linewidth]{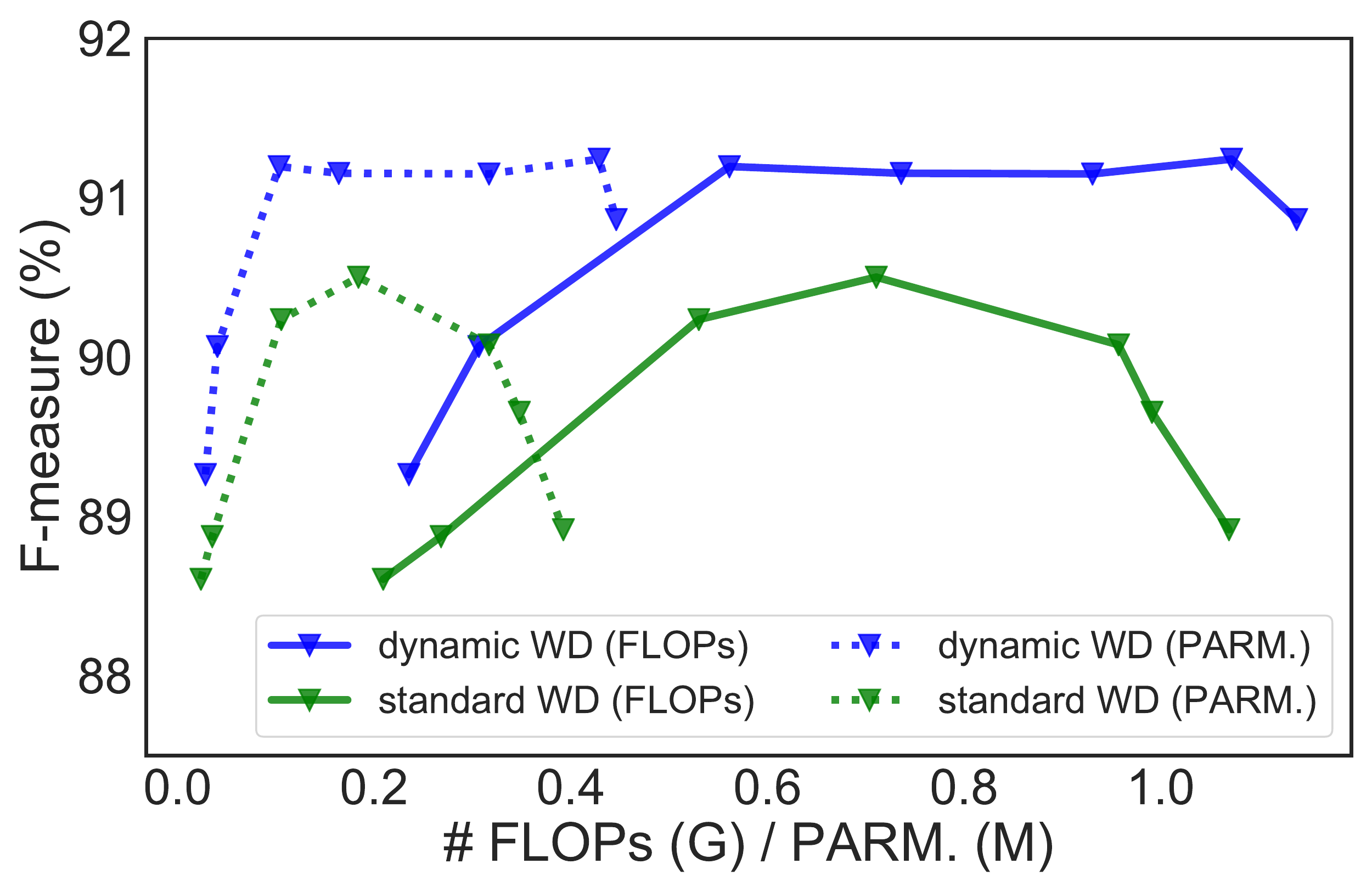}
  \centering
  \end{minipage}
  \begin{minipage}[t]{0.49\linewidth}
  \centering
  \includegraphics[width=\linewidth]{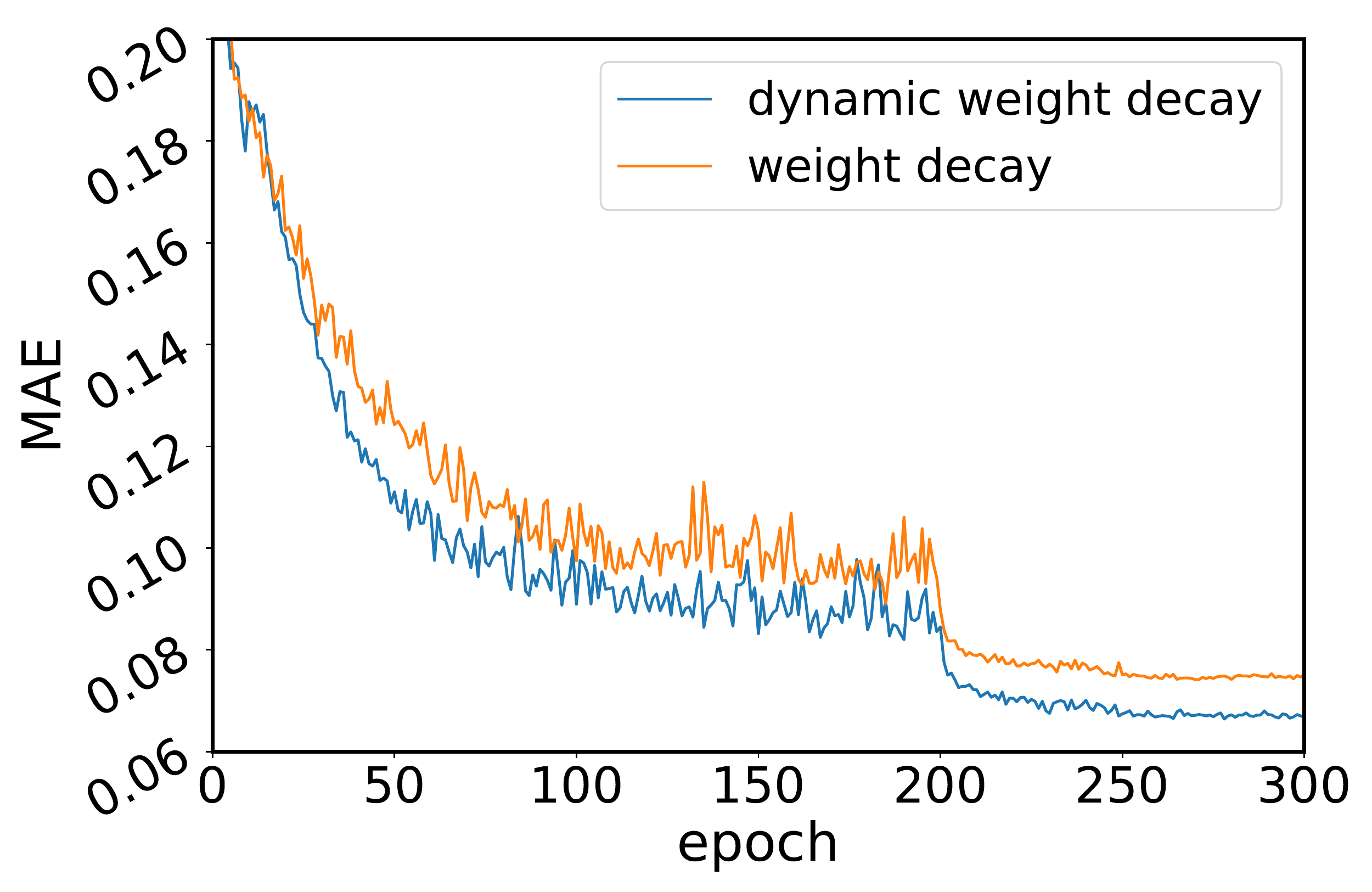}
  \end{minipage}
   \caption{a) Left: Performance and complexity of compressed model using dynamic/standard weight decay under 
    different $\lambda$ as shown in \eqref{eq:weight_decay}. 
     b) Right: The test MAE of models with/without dynamic weight decay.}
    \label{fig:performance_wd_trade_off_gap_mae}
\end{figure}

\subsection{Dynamic Weight Decay}
\label{sec:dynamic_wd_exp}

In this section, we verify the effectiveness of our proposed dynamic weight decay.
We apply different degrees of standard weight decay to achieve the trade-off between model performance and sparsity, while keeping the weights for dynamic weight decay unchanged.
We insert our proposed dynamic weight decay to the weights of BatchNorm layers
while using the standard weight decay on remaining weights for a fair comparison.
\figref{fig:performance_wd_trade_off_gap_mae} (a) shows the performance and complexity of the compressed model using dynamic/standard weight decay under different $\lambda$ in \eqref{eq:weight_decay}.
Models trained with dynamic weight decay have better performance under the same complexity.
Also, the performance of dynamic weight decay based models is less sensitive to the model complexity.
We eliminate redundant channels according to the absolute value of $\gamma$ in \eqref{eq:BatchNorm}
as described in \secref{sec:search_split_ratio_theory}.
\figref{fig:gap_std_bn_weight_distribution} (b) shows the distribution of $\gamma$ for models trained with/without dynamic weight decay.
By suppressing weights according to features, 
dynamic weight decay enforces the model with more sparsity.
\figref{fig:gap_std_bn_weight_distribution} (a) reveals the average standard deviation of outputs among channels
from the BatchNorm layer of models trained with/without dynamic weight decay.
Features of dynamic weight decay based model are more stabilized
due to the stable weight distribution.
\figref{fig:performance_wd_trade_off_gap_mae} (b) shows the testing MAE of each epoch
with/without dynamic weight decay.
Training with dynamic weight decay brings
better performance in terms of MAE and faster convergence. 
The dynamic weight decay generalizes well on other tasks such as classification, and semantic segmentation.
\begin{table}[t]
  \centering
  \begin{tabular}{lcccc|lcccc}\toprule
    \multicolumn{5}{l}{Pruning Filters~\cite{li2016pruning}} &   \multicolumn{3}{l}{Geometric-Median~\cite{he2019filter}}\\ \midrule
                & PARM.   & FLOPs  & $F_\beta$  & $M$        &                   & PARM.   & FLOPs  & $F_\beta$  & $M$    \\ \midrule
  Standard&227K&0.69G&88.7&0.080 &Standard&227K&0.70G&88.7&0.083 \\  
  Dynamic &226K&0.69G&89.4&0.078 &Dynamic &226K&0.68G&89.6&0.082  \\ 
  \bottomrule 
  \end{tabular}
  \tabSpace
  \caption{Integrating dynamic weight decay into pruning methods.
  Standard/Dynamic denote the standard/dynamic weight decay, respectively.
  }
  \label{tab:cooperating_pruning}
\end{table}

\noindent\textbf{Cooperating with pruning methods.}
By default, we use the pruning method in~\cite{liu2017learning}
to eliminate redundant weights.
Since our proposed dynamic weight decay focuses on introducing
sparsity while maintaining a stable and compact distribution of weights among channels,
it is orthogonal to commonly use pruning methods that focus
on identify unnecessarily weights.
Therefore, we integrate the dynamic weight decay into several pruning methods as shown in \tabref{tab:cooperating_pruning}.
All configurations remain the same except for replacing the standard weight decay to our proposed dynamic weight decay.
Pruning methods~\cite{li2016pruning,he2019filter} equipped with dynamic weight decay achieve
better performance under the similar parameters.

\newcommand{\mRows}[1]{\multirow{2}{*}{#1}}

\begin{table}[t]
\tabFormat
\setlength{\tabcolsep}{1.4mm}
\begin{tabular}{llccccc}\toprule
\multicolumn{1}{l}{Width} & Prune & $\times$1 & $\times$1.2 & $\times$1.5 & $\times$1.8 & $\times$2.0 \\  \midrule
\mRows{Parms} & N & 211K & 298K & 455K & 645K & 788K \\ 
              & Y &  94K & 109K & 118K & 134K & 140K \\ \midrule
Ratio         &  & 55$\%$ &63$\%$ &74$\%$ &79$\%$ &82$\%$ \\ \midrule
\mRows{FLOPs} & N & 0.61G & 0.82G & 1.17G & 1.58G & 1.87G \\ 
              & Y & 0.43G & 0.52G & 0.63G & 0.71G & 0.72G \\ \midrule 
Ratio         &   &30$\%$ &37$\%$ &46$\%$ &55$\%$ &61$\%$ \\ \midrule
\mRows{$F_\beta$}&N & 90.0  & 90.7 & 91.1 & 91.2 & 91.5 \\
              & Y & 90.0 & 90.7 & 91.2 &  91.3 & 91.6 \\
\bottomrule 
\end{tabular}
\tabSpace
\caption{The compression ratio of CSNet with different initial channel widths.
The pruning rate is defined as the ratio of model complexity 
between pruned parts and complete CSNet.}
\label{tab:compression_width}
\end{table}

\subsection{CSNet with Learned Channels in gOctConv} \label{sec:searched_model}

\noindent\textbf{Pruning rate $\&$ Channel width.} 
An initial training space with a large channel width is required
for learning more useful features.
To enlarge the available training space,
we linearly expand the channel number of gOctConvs.
A pruning rate is defined as the ratio of model complexity 
between pruned parts and complete CSNet.
\tabref{tab:compression_width} shows the pruning rate of CSNet with different initial channel widths. 
The split-ratio of gOctConvs for the initial model is set to 5/5.
Larger initial width results in better performance as expected.
As the initial width rises, the complexity of pruned models
only has a limited increment. 
The quality of the pruned model is dependant on the available training space.
With a large enough training space, results are closing to the optimal.
Also, benefited from the stable distribution introduced by dynamic weight decay,
compressed models have similar or even better performance compared with the initial model.
\begin{figure}[t]
  \centering
  \begin{overpic}[width=0.6\linewidth]{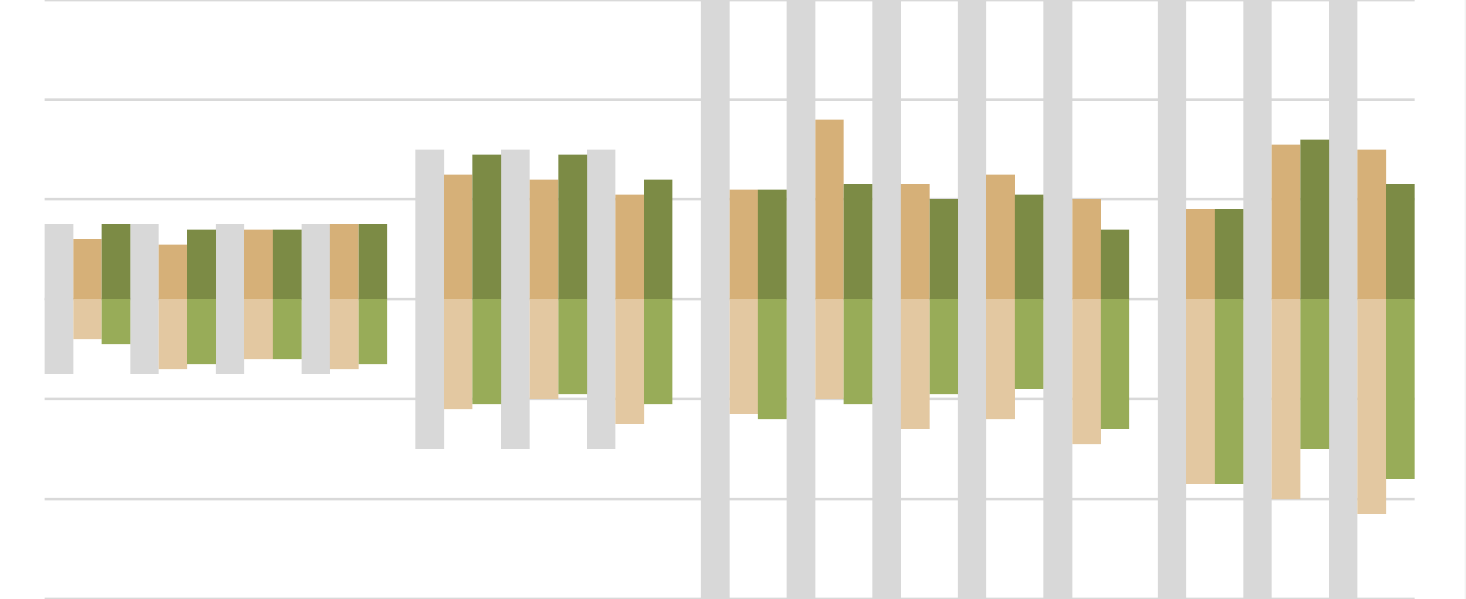}
  \put(2, 35){High resolution}
  \put(2, 1){Low resolution}
  \put(98, 24.5){\rotatebox{90}{channels}}
  \end{overpic}
  \caption{Visualization of the feature extractor of CSNet.
    Gray is CSNet with fixed channel, Yellow and Green are the CSNet-L trained with standard/dynamic weight decay, respectively.} 
  \label{fig:histgrams-prune}
\end{figure}

\noindent\textbf{Visualization of channels of gOctConvs.} We visualize the learned channel number of gOctConvs in \figref{fig:histgrams-prune}.
It can be seen that as the network goes deeper, the feature extraction network
shows a trend of utilizing more low resolution features.
Within the same stage, high resolution features are urged in the middle of the stage.
Also, the model trained with dynamic weight decay has a stabler
channel number variation among different layers.
Deeper layers contain more redundant channels compared with shallow layers.  
\subsection{Run-Time}
We compare the run-time of our proposed CSNet with existing models from~\tabref{tab:performance_sota} as shown in~\tabref{tab:run-time}.
The run-time is tested with an image of 224 $\times$ 224 on a single core of i7-8700K CPU.
Our proposed CSNet has more than x10 acceleration compared with large-weight models.
With similar speed, CSNet achieves up to $6\%$ gain in F-measure compared with those models designed for other tasks.
However, there is still a gap between FLOPs and run-time,
as current deep learning frameworks are not optimized for vanilla and our proposed gOctConvs.
\begin{table}[t]
  \tabFormat
  \setlength{\tabcolsep}{1.2mm}
  \begin{tabular}{lcc|lcc}\toprule
    Method                       & FLOPs        & Run-time       & Method                       & FLOPs         & Run-time      \\  \midrule
    PiCANet~\cite{Liu_2016_CVPR} & 54.06G       & 2850.2ms       & PoolNet~\cite{Liu19PoolNet}  & 88.89G        & 997.3ms      \\ 
    ENet~\cite{paszke2016enet}   & 0.40G        & 89.9ms         & ESPNetv2~\cite{mehta2019espnetv2} & 0.31G    & 186.3ms      \\ 
    CSNet$\times$1               &  0.61G       & 135.9ms        & CSNet$\times$1-L                  & 0.43G    & 95.3ms       \\     
  \bottomrule 
  \end{tabular}
  \tabSpace
  \caption{Run-time using 224 $\times$ 224 input on a single core of i7-8700K CPU.}
  \label{tab:run-time}
  \end{table}


\section{Conclusion}
In this paper, we propose the generalized OctConv with
more flexibility
to  efficiently utilize both in-stage and cross-stages multi-scale features, 
while reducing the representation redundancy by a novel dynamic weight decay scheme.
The dynamic weight decay scheme maintains a stable weights distribution among channels and stably boosts the sparsity of parameters during training.
Dynamic weight decay supports learnable number of channels for each scale in gOctConvs, 
allowing $80\%$ of parameters reduce with negligible performance drop.
Utilizing different instances of gOctConvs, we build an extremely light-weighted model, namely CSNet,  which  achieves  comparable  performance  with $\sim0.2\%$  parameters (100k) of large models on popular salient object detection benchmarks.

\myPara{\textbf{Acknowledgements}} 
Ming-Ming Cheng is the corresponding author.
This research was supported by Major Project for New
Generation of AI under Grant No. 2018AAA0100400,
NSFC (61922046), Tianjin Natural Science Foundation (18ZXZNGX00110),
and the Fundamental Research Funds for the Central Universities,
Nankai University (63201169).

%
%
\bibliographystyle{splncs04}
\bibliography{ref}
\end{document}